\documentclass{article}
\pdfoutput=1
\usepackage{odyssey_04,amssymb,amsmath,epsfig}

\usepackage{multirow} 
\usepackage{array} 
\usepackage{subfig}
\usepackage{url} 

\renewcommand{\tablename}{Table~} 
\renewcommand{\figurename}{Figure~} 

\ninept

\title{Can Facial Uniqueness be Inferred from Impostor Scores?}

\makeatletter
\def\name#1{\gdef\@name{#1\\}}
\makeatother
\name{{\em Abhishek Dutta, Raymond Veldhuis, Luuk Spreeuwers}
  \thanks{This work was supported by the BBfor2 project which is funded by the EC as a Marie-Curie ITN-project (FP7-PEOPLE-ITN-2008) under Grant Agreement number 238803.}}

\address{University of Twente, Netherlands  \\
{\small \tt \{a.dutta,r.n.j.veldhuis,l.j.spreeuwers\}@utwente.nl} }
\begin{document}
\maketitle
\begin{abstract}
In Biometrics, facial uniqueness is commonly inferred from impostor similarity scores. In this paper, we show that such uniqueness measures are highly unstable in the presence of image quality variations like pose, noise and blur. We also experimentally demonstrate the instability of a recently introduced impostor-based uniqueness measure of [Klare and Jain 2013] when subject to poor quality facial images.
\end{abstract}

\section{Introduction}
The appearances of some human faces are more similar to facial appearances of other subjects in a population. Those faces whose appearance is very different from the population are often called a unique face. Facial uniqueness is a measure of distinctness of a face with respect to the appearance of other faces in a population. Non-unique faces are known to be more difficult to recognize by the human visual system \cite{going1974effects}  and automatic face recognition systems \cite[Fig.~6]{klare2012face}. Therefore, in Biometrics, researchers have been actively involved in measuring uniqueness from facial photographs \cite{klare2012face,ross2009exploiting,yager2010biometric,wittman2006empirical}. Such facial uniqueness measurements are useful to build an adaptive face recognition system that can apply stricter decision thresholds for fairly non-unique facial images which are much harder to recognize.

Most facial uniqueness measurement algorithms quantify the uniqueness of a face by analyzing its similarity score (i.e.\ impostor score) with the facial image of other subjects in a population. For example, \cite{klare2012face} argue that a non-unique facial image (i.e.\ lamb\footnote{sheep: easy to distinguish given a good quality sample, goats: have traits difficult to match, lambs: exhibit high levels of similarity to other subjects, wolves: can best mimic other subject's traits} as defined in \cite{doddington1998sheep}) ``will generally exhibit high level of similarity to many other subjects in a large population (by definition)''. Therefore, they claim that facial uniqueness of a subject can be inferred from its impostor similarity score distribution.

In this paper, we show that impostor scores are not only influenced by facial identity (which in turn defines facial uniqueness) but also by quality aspects of facial images like pose, noise and blur. Therefore, we argue that a facial uniqueness measure based solely on impostor scores may give misleading results for facial images degraded by quality variations.

The organization of this paper is as follows: in Section \ref{sec:related_work}, we review some existing methods that use impostor scores to measure facial uniqueness, next in Section \ref{sec:quality_influence_imp_score} we describe the experimental setup that we use to study the influence of facial identity and image quality on impostor scores, in Section \ref{sec:ium_stability} we investigate the stability of one recently introduced impostor-based uniqueness measure (i.e.\ \cite{klare2012face}). Finally, in Section \ref{sec:discussion}, we discuss the experimental results and present the conclusions of this study in Section \ref{sec:conclusion}.

\section{Related Work}
\label{sec:related_work}
Impostor score distribution has been widely used to identify the subjects that exhibit high level of similarity to other subjects in a population (i.e.\ lamb). The authors of \cite{doddington1998sheep} investigated the existence of ``lamb'' in speech data by analyzing the relative difference between maximum impostor score and genuine score of a subject. They expected the ``lambs'' to have very high maximum impostor score. A similar strategy was applied by \cite{wittman2006empirical} to locate non-unique faces in a facial image dataset. The authors of \cite{ross2009exploiting} tag a subject as ``lamb'' if its mean impostor score lies above a certain threshold. Based on this knowledge of a subject's location in the ``Doddington zoo'' \cite{doddington1998sheep}, they propose an adaptive fusion scheme for a multi-modal biometric system. Recently, \cite{klare2012face} have proposed an Impostor-based Uniqueness Measure (IUM) which is based on the location of mean impostor score relative to the maximum and minimum of the impostor score distribution. Using both genuine and impostor scores, \cite{yager2010biometric} investigated the existence of biometric menagerie in a broad range of biometric modalities like 2D and 3D faces, fingerprint, iris, speech, etc.

All of these methods that aim to measure facial uniqueness from impostor scores assume that impostor score is only influenced by facial identity. In this paper, we show that impostor scores are also influenced by image quality (like pose, noise, blur, etc). 

The authors of \cite{paone2011difficult} have also concluded that facial uniqueness (i.e.\ location in the biometric zoo) changes easily when imaging conditions (like illumination) change. Their conclusion was based on results from a single face recognition system (i.e. FaceVACS \cite{facevacs2010}). In this paper, we also investigate the characteristics of facial uniqueness using four face recognition systems (two commercial and two open-source systems) operating on facial images containing the following three types of quality variations: pose, blur and noise.


\section{Influence of Image Quality on Impostor Score Distribution}
\label{sec:quality_influence_imp_score}
\begin{figure}[ht]
 \centering
 \includegraphics[width=0.4\linewidth]{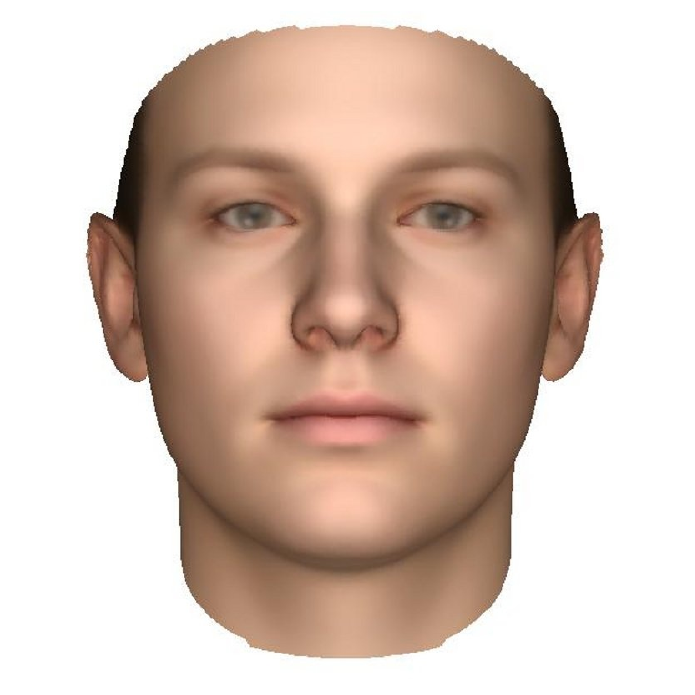}
 \caption{Average face image}
 \label{fig:bfm_average_face}
\end{figure}

In this section, we describe an experimental setup to study the influence of image quality on impostor scores. We fix the identity of query image to an average face image synthesized\footnote{using the code and model provided with \cite{paysan20093dface}} by setting the shape ($\alpha$) and texture ($\beta$) coefficients to zero $(\alpha, \beta = 0)$ as shown in \figurename\ref{fig:bfm_average_face}. We obtain a baseline impostor score distribution by comparing the similarity between the average face and a gallery set (or, impostor population) containing $250$ subjects. Now, we vary the quality (pose, noise and blur) of this gallery set (identity remains fixed) and study the variation of impostor score distribution with respect to the baseline. Such a study will clearly show the influence of image quality on impostor score distribution as only image quality varies while the facial identity remains constant in all the experiments.


We use the MultiPIE neutral expression dataset of \cite{gross2008multipie} to create our gallery set. Out of the 337 subjects in MultiPIE, we select 250 subjects that are common in session (01,03) and session (02,04). In other words, our impostor set contains subjects from $(S_1 \cup S_3) \cap (S_2 \cup S_4)$, where $S_i$ denotes the set of subjects in MultiPIE session $i \in \{1,2,3,4\}$ recording $1$. From the group $(S_1 \cup S_3)$, we have $407$ images of $250$ subject and from the group $(S_2 \cup S_4)$, we have $413$ images of the same $250$ subjects. Therefore, for each experiment instance, we have $820$ images of $250$ subjects with at least two image per subject taken from different sessions.

We compute the impostor score distribution using the following four face recognition systems: FaceVACS \cite{facevacs2010}, Verilook \cite{verilook2011}, Local Region PCA and Cohort LDA \cite{bolme2012csu}. The first two are commericial while the latter two are open source face recognition systems. We supply the same manually labeled eye coordinates to all the four face recognition systems in order to avoid the performance variation caused by automatic eye detection error.

\begin{figure}[ht]
 \centering
 \includegraphics[width=\linewidth]{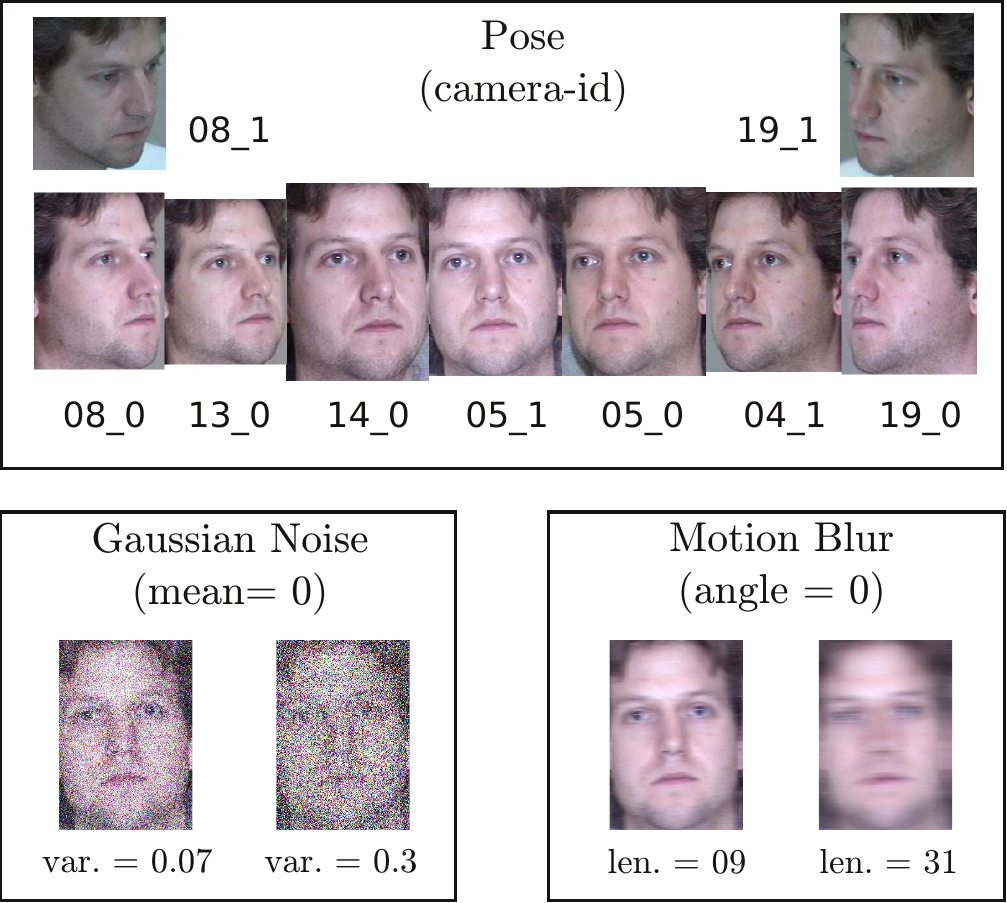}
 \caption{Facial image quality variations included in this study.}
 \label{fig:qual_var_illus}
\end{figure}

In this experiment, we consider impostor population images with frontal view (cam $05\_1$) and frontal illumination (flash $07$) images as the baseline quality. We consider the following three types of image quality variations of the impostor population: pose, blur, and noise as shown in \figurename\ref{fig:qual_var_illus}. For pose, we vary the camera-id (with flash that is frontal with respect to the camera) of the impostor population. For noise and blur, we add artificial noise and blur to frontal view images (cam $05\_1$) of the impostor population. We simulate imaging noise by adding zero mean Gaussian noise with the following variances: $\{0.007, 0.03, 0.07, 0.1, 0.3\}$ (where pixel value is in the range $[0,1.0]$). To simulate $N$ pixel horizontal linear motion of subject, we convolve frontal view images with a $1 \times N$ averaging filter, where $N \in \{3,5,7,13,17,29,31\}$ (using Matlab's \texttt{fspecial('motion', N, 0)} function). For pose variation, camera-id $19\_1$ and $08\_1$ refer to right and left surveillance view images respectively. 

In \figurename\ref{fig:fv_csu_qri_rf0}, we report the variation of impostor score distribution of the average face image as box plots \cite{wickham2009ggplot2}. In these box plot, the upper and lower hinges correspond to the first and third quantiles. The upper (and lower) whisker extends from the hinge to the highest (lowest) value that is within $1.5 \times $IQR where IQR is the distance between the first and third quartiles. The outliers are plotted as points. 

\section{Stability of Impostor-based Uniqueness Measure Under Quality Variation}
\label{sec:ium_stability}
In this section, we investigate the stability of a recently proposed impostor-based facial uniqueness measure \cite{klare2012face} under image quality variations. The key idea underpinning this method is that a fairly unique facial appearance will result in low similarity score with a majority of facial images in the population. This definition of facial uniqueness is based on the assumption that similarity score is influenced only by facial identity. 

This facial uniqueness measure is computed as follows: Let $i$ be a probe (or query) image and $J=\{j_1,\cdots,j_n\}$ be a set of facial images of $n$ different subjects such that $J$ does not contain an image of the subject present in image $i$. In other words, $J$ is the set of impostor subjects with respect to the subject in image $i$. If $S = \{s(i,j_1), \cdots, s(i,j_n)\}$ is the set of similarity score between image $i$ and the set of images in $J$, then the Impostor-based Uniqueness Measure (IUM) is defined as:
\begin{equation}
  u(i,J) = \frac{S_{max} - \mu_{S}}{S_{max} - S_{min}}
 \label{eq:ium_score_computation}
\end{equation}
where,  $S_{min}, S_{max}, \mu_{S}$ denote minimum, maximum and average value of impostor scores in $S$ respectively. A facial image $i$ which has high similarity with a large number of subjects in the population will have a small IUM value $u$ while an image containing highly unique facial appearance will take a higher IUM value $u$.

\begin{figure*}[ht]
 \centering
 \includegraphics[width=0.9\linewidth]{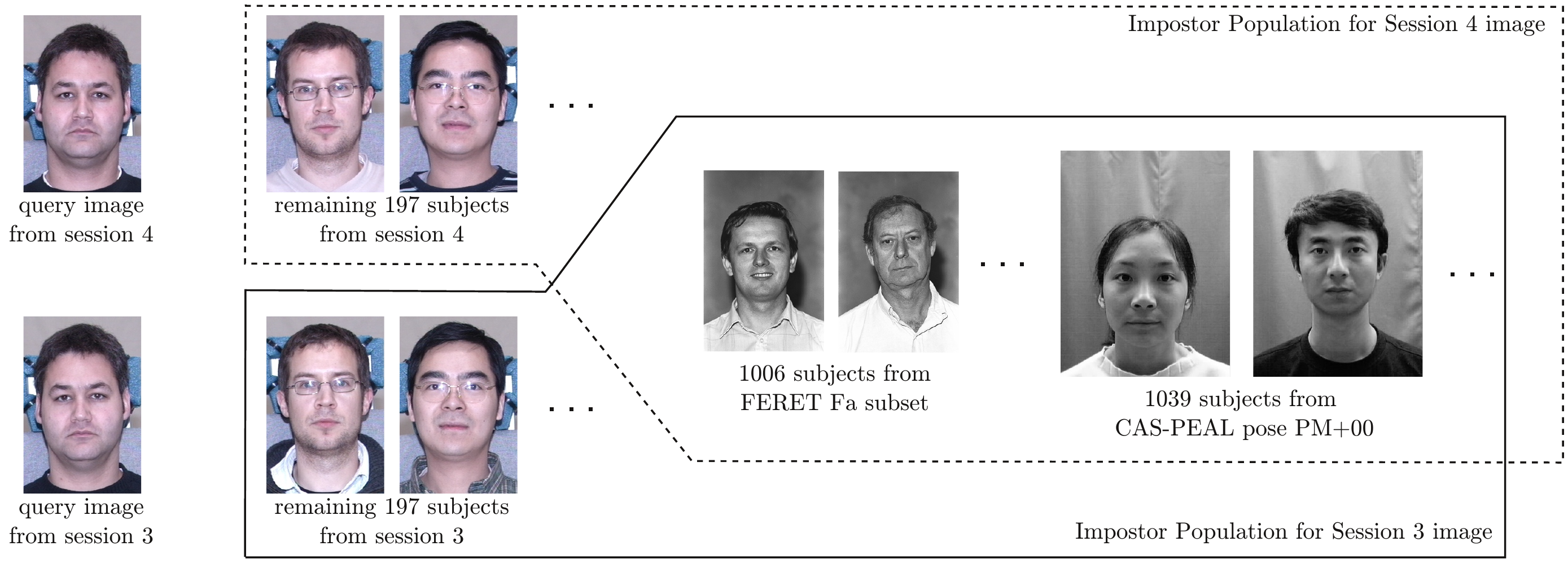}
 \caption{Selection of impostor population for IUM score computation.}
 \label{fig:klare2012face_exp_data_illus}
\end{figure*}

For this experiment, we compute the IUM score of $198$ subjects common in session $3$ and $4$ (i.e.\ $S_3 \cap S_4$) of the MultiPIE dataset. The IUM score corresponding to same identity but computed from two different sessions (the frontal view images without any artificial noise or blur) must be highly correlated. We denote this set of IUM scores as the baseline uniqueness scores. To study the influence of image quality on the IUM scores, we only vary the quality (pose, noise, blur as shown in  \figurename\ref{fig:qual_var_illus}) of the session $4$ images and we compute the IUM scores under quality variation. If the IUM scores are stable with image quality variations, the IUM scores computed from session $3$ and $4$ should remain highly correlated despite quality variation in session $4$ images. Recall that the facial identity remains fixed to the same $198$ subjects in all these experiments.

In \cite{klare2012face}, the authors compute IUM scores from an impostor population of $(16000-1)$ subjects taken from a private dataset. We do not have access to such a large dataset. Therefore, we import additional impostors from CAS-PEAL dataset ($1039$\footnote{in our version of the CAS-PEAL dataset, PM+00 images for person-id $261$ in the pose subset were missing. Therefore, we use only 1039 of the total 1040 subjects in the original dataset} subjects from pose subset PM+00) \cite{gao2008cas} and FERET (1006 subjects from Fa subset) \cite{phillips2000feret}. So, for computing the IUM score for subject $i$ in session $3$, we have a impostor population containing the remaining 197 subjects from session $3$, $1039$ subjects from CAS-PEAL and $1006$ subjects from FERET. Therefore, each of the IUM score is computed from an impostor set $J$ containing a single frontal view images of $197+1039+1006=2242$ subjects as shown in \figurename\ref{fig:klare2012face_exp_data_illus}. In a similar way, we compute IUM scores for the same $198$ subjects but with images taken from session $4$. As the Cohort LDA system requires colour images, we replicate the gray scale images of FERET and CAS-PEAL in RGB channels to form a colour image. Note that we only vary the quality of a single query facial image $i$ (from session $4$) while keeping the impostor population quality $J$ fixed to $2242$ frontal view images (without any artificial noise or blur).

In \tablename\ref{tbl:session_3_4_ium_correlation}, we show the variation of Pearson correlation coefficient (\texttt{cor()} \cite{r-project}) between IUM scores of $198$ subjects computed from session $3$ and $4$. The bold faced entries correspond to the correlation between IUM scores computed from frontal view (without any artificial noise or blur) images of the two sessions. The remaining entries denote variation in correlation coefficient when the quality of facial image in session $4$ is varied without changing the quality of impostor set. In \figurename\ref{fig:norm_corr_coef_ium}, we show the drop-off of normalized correlation coefficient (derived from \tablename\ref{tbl:session_3_4_ium_correlation}) with quality degradation where normalization is done using baseline correlation coefficient.

\section{Discussion}
\label{sec:discussion}

\subsection{Influence of Image Quality on Impostor Score}
\begin{figure*}[ht]
 \centering
 \includegraphics[width=\linewidth]{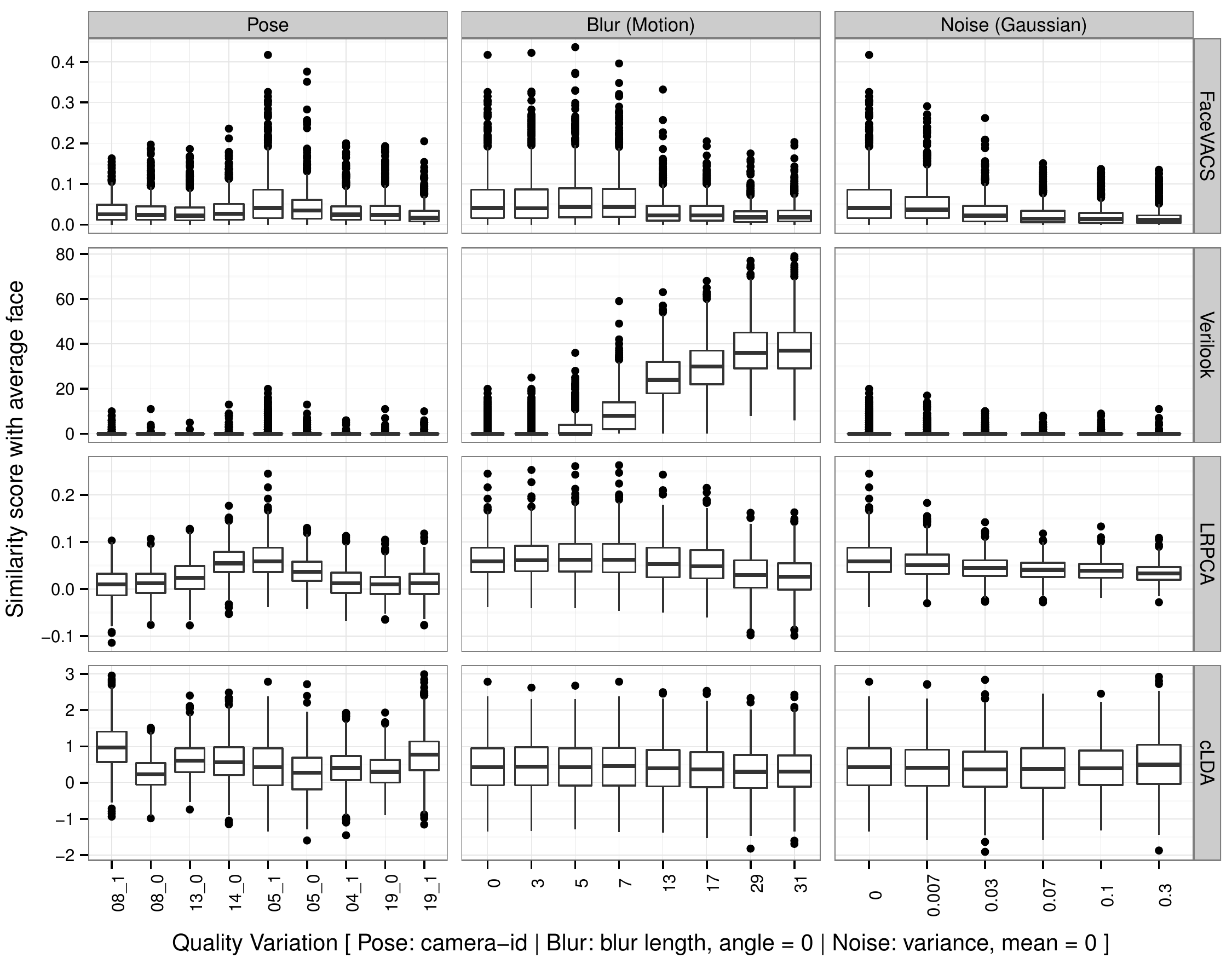}
 \caption{Influence of image quality on impostor score distribution shown as box plot where the outliers are plotted as points. The rows correspond to a particular face recognition system (FaceVACS, Verilook, LRPCA, cLDA) and the columns correspond to the following three image quality variations: pose, motion blur and Gaussian noise.}
 \label{fig:fv_csu_qri_rf0}
\end{figure*}

In \figurename\ref{fig:fv_csu_qri_rf0}, we show the variation of impostor score distribution with image quality variations of the impostor population. We consider frontal view (cam $05\_1$) image without any artificial noise or blur (i.e.\ the original image in the dataset) as the baseline image quality. The box plot corresponding to cam-id=$05\_1$, blur-length=$0$, noise-variance=$0$ denotes mainly the impostor score variation due to facial identity. 

In \figurename\ref{fig:fv_csu_qri_rf0}, we observe that, the nature of impostor score distribution corresponding to all three types of quality variations is significantly different from the baseline impostor distribution. For instance, the impostor score distribution for FaceVACS and Verilook systems corresponding to a motion blur of length $31$ pixels is completely different from that corresponding to no motion blur. Furthermore, the impostor score distribution also seem to be responding to quality variations. For example, the mean of impostor distribution for FaceVACS system appears to increase monotionically as the image quality moves towards the baseline image quality. We also observe that the impostor score distribution of the four face recognition systems respond in a different way to the three types of image quality variations. These observations clearly show that the impostor score distribution is not only influenced by identity (as expected) but also by the image quality like pose, blur and noise.

\subsection{Stability of Impostor-based Uniqueness Measure Under Quality Variation}
We observe a common trend in the variation of correlation coefficients with image quality degradation as shown in \tablename\ref{tbl:session_3_4_ium_correlation}. The correlation coefficient is maximum for the baseline image quality (frontal, no artificial noise or blur). As we move away from the baseline image quality, the correlation between IUM scores reduces. This reduction in correlation coefficient indicates the instability of Impostor-based Uniqueness Measure (IUM) in the presence of image quality variations.

\begin{figure*}[ht]
 \centering
 \includegraphics[width=0.8\linewidth]{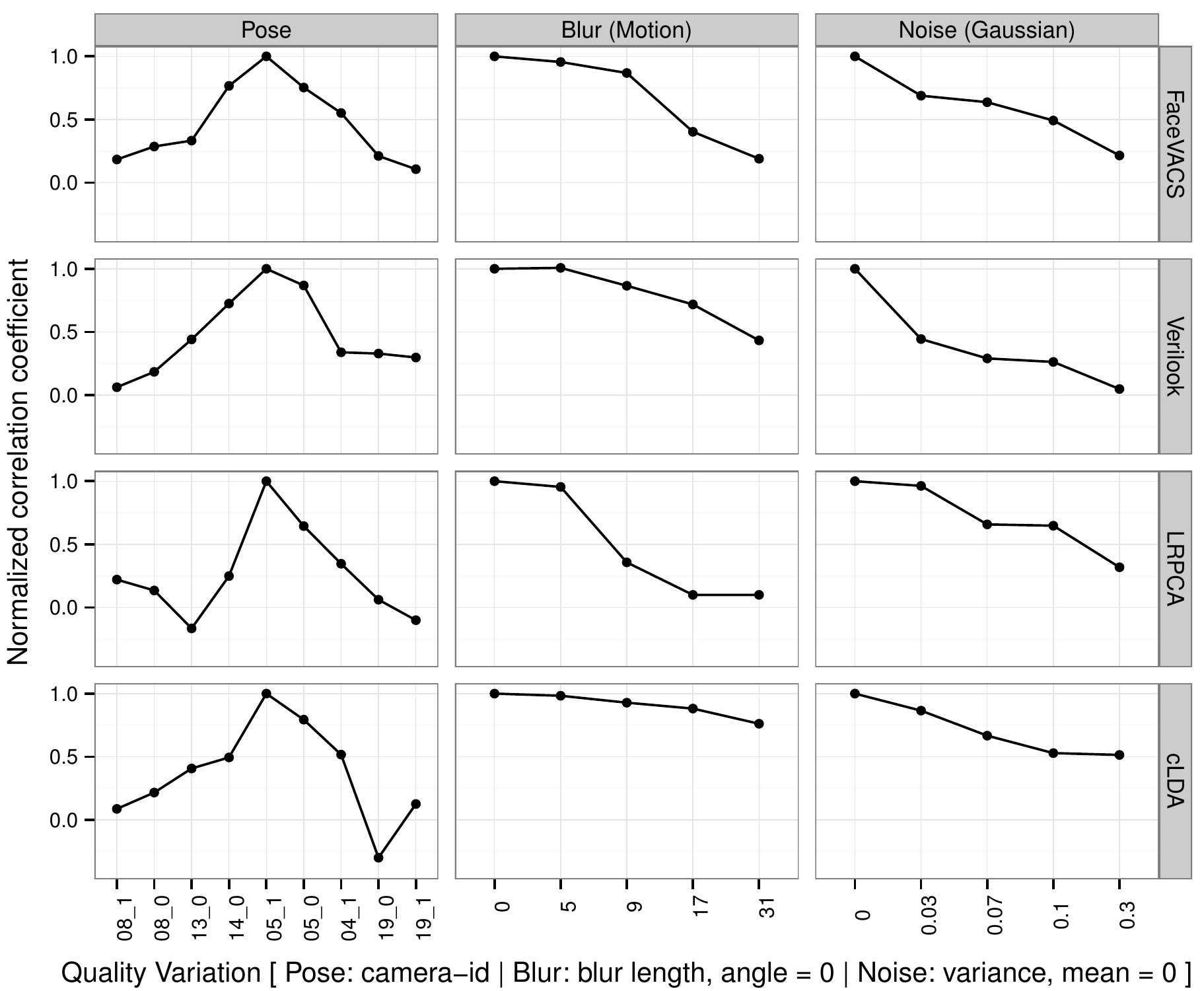}
 \caption{Fall-off of normalized correlation coefficient with quality degradation. Normalization performed using correlation coefficient corresponding to frontal, no blur and no noise case.}
 \label{fig:norm_corr_coef_ium}
\end{figure*}

The instability of IUM is also depicted by the normalized correlation coefficient plot of \figurename\ref{fig:norm_corr_coef_ium}. For all the four face recognition systems, we observe fall-off of the correlation between IUM scores with variation in pose, noise and blur of facial images. For pose variation, peak correlation is observed for frontal view (camera 05\_1) facial images because, in this case, both pairs of IUM scores correspond to frontal view images taken from two session $3$ and session $4$. 

The instability of IUM measure is also partly due to the use of minimum and maximum impostor scores in equation (\ref{eq:ium_score_computation}) which makes it more susceptible to outliers.

The authors of \cite{klare2012face}, who originially proposed the Impostor-based Uniqueness measure (IUM), report a correlation of~$\geq~0.92$ using FaceVACS system on a privately held mug shot database of 16000 subjects created from the operational database maintained by the Pinellas County Sheriff's Office. Futher details about the quality of facial images in this dataset is not available. From the sample images shown in \cite{klare2012face}, we can assume that this private mugshot database contains sharp frontal view facial images captured under uniform illumination. Our baseline image quality (frontal view without any artificial blur or noise) comes very close to the quality of images used in their experiment. However, we get a much lower correlation coefficient of $\leq 0.68$ on a combination of three publicly released dataset. One reason for this drop in correlation may be due to difference in the quality (like resolution) of facial images. Our impostor population is formed using images taken from three publicly available dataset and therefore represents larger variation in image quality as shown in \figurename\ref{fig:klare2012face_exp_data_illus}. To a lesser extent, this difference in correlation could also be due to difference in the FaceVACS SDK version used in the two experiments. We use the FaceVACS SDK version 8.4.0 (2010) and they have not mentioned the SDK version used in their experiments.

\section{Conclusion}
\label{sec:conclusion}
We have shown that impostor score is influenced by both identity and quality of facial images. We have also shown that any attempt to measure characteristics of facial identity (like facial uniqueness) solely from impostor score distribution shape may give misleading results in the presence of image quality degradation in the input facial images.

This research has thrown up many questions in need of further investigation regarding the stability of existing facial uniqueness measures based solely on impostor scores. More research is needed to better understand the impact of image quality on the impostor score distribution. Such studies will help develop uniqueness measures that are robust to quality variations.

\section{Acknowledgement}
\begin{itemize}
 \item We would like to thank Cognitec Systems GmbH. for supporting our research by providing the FaceVACS software. Results obtained for FaceVACS were produced in experiments conducted by the University of Twente, and should therefore not be construed as a vendor's maximum effort full capability result.
 \item We also acknowledge the anonymous reviewers of the  BTFS 2013 conference for their valuable feedback.
\end{itemize}

%
%

\begin{table*}[ht]
 \centering
 \subfloat{
 \centering
 \begin{tabular}{l|c|c|c|c|>{\bfseries}c|c|c|c|c}
 \hline
 & 08\_1 & 08\_0 & 13\_0 & 14\_0 & frontal & 05\_0 & 04\_1 & 19\_0 & 19\_1 \\
\hline
FaceVACS & 0.12 & 0.19 & 0.23 & 0.52 & 0.68 & 0.51 & 0.37 & 0.14 & 0.07 \\
Verilook & 0.04 & 0.12 & 0.28 & 0.45 & 0.63 & 0.54 & 0.21 & 0.21 & 0.19 \\
LRPCA & 0.10 & 0.06 & -0.07 & 0.11 & 0.45 & 0.29 & 0.15 & 0.03 & -0.05 \\
cLDA & 0.04 & 0.09 & 0.17 & 0.21 & 0.43 & 0.34 & 0.22 & -0.13 & 0.05 \\
\hline
 \multicolumn{5}{r|}{$\xleftarrow{\textrm{drop in correlation with pose}}$} & baseline & \multicolumn{4}{l}{$\xrightarrow{\textrm{drop in correlation with pose}}$} \\

 \end{tabular}
 }

 \subfloat{
 \begin{tabular}{l|>{\bfseries}c|c|c|c|c}
 \hline
& No blur & length 5 & length 9 & length 17 & length 31 \\
 \hline
FaceVACS & 0.68 & 0.65 & 0.59 & 0.27 & 0.13 \\
Verilook & 0.63 & 0.63 & 0.54 & 0.45 & 0.27 \\
LRPCA & 0.45 & 0.43 & 0.16 & 0.04 & 0.04 \\
cLDA & 0.43 & 0.42 & 0.40 & 0.38 & 0.32 \\
\hline
 & baseline & \multicolumn{4}{l}{$\xrightarrow{\textrm{drop in correlation with blur}}$} \\
 \end{tabular}
 }

 \subfloat{
 \begin{tabular}{l|>{\bfseries}c|c|c|c|c}
 \hline
 & No noise & $\sigma=0.03$ &  $\sigma=0.07$ &  $\sigma=0.1$ &  $\sigma=0.3$ \\
 \hline
FaceVACS & 0.68 & 0.47 & 0.43 & 0.33 & 0.15 \\
Verilook & 0.63 & 0.28 & 0.18 & 0.16 & 0.03 \\
LRPCA & 0.45 & 0.43 & 0.29 & 0.29 & 0.14 \\
cLDA & 0.43 & 0.37 & 0.28 & 0.23 & 0.22 \\
\hline
 & baseline & \multicolumn{4}{l}{$\xrightarrow{\textrm{drop in correlation with noise}}$} \\
 \end{tabular}
 }
 \caption{Variation in correlation of the impostor-based uniqueness measure \cite{klare2012face} for $198$ subjects computed from sessions $3$ and $4$. Note that image quality (pose, noise and blur) of session $4$ images were only varied while session $3$ and impostor population images were fixed to frontal view images without any artificial noise or blur.}
 \label{tbl:session_3_4_ium_correlation}
\end{table*}

\bibliographystyle{IEEEbib}
\bibliography{/home/abhishek/sas-pc-56-DISK2_400G/UT/phd/research/phd_thesis/tex_src/references/phd_references}

\end{document}